%% file: main.tex
\documentclass{article}
\usepackage{spconf,amsmath,graphicx}
\usepackage[utf8]{inputenc}
\usepackage{amsmath,amssymb}
\usepackage{booktabs}
\usepackage{color}
\usepackage{framed}
\usepackage{graphics}
\usepackage{graphicx}
\usepackage{placeins}
\usepackage{soul}
\usepackage{url}
\usepackage{cleveref}
\usepackage{multirow}
\usepackage{flushend}

\makeatother
\definecolor{darkgreen}{RGB}{0, 140, 0}
\definecolor{orange}{RGB}{250, 140, 0}

\newcommand{\etal}{\emph{et al.}}

\makeatletter
\renewcommand{\paragraph}{%
  \@startsection{paragraph}{4}%
  {\z@}{0.4em}{-1em}%
  {\normalfont\normalsize\bfseries}%
}

\title{Fixing the train-test resolution discrepancy: FixEfficientNet}
\name{Hugo Touvron, Andrea Vedaldi, Matthijs Douze, Herv\'e J\'egou}
\address{Facebook AI Research}

\begin{document}

\maketitle
\begin{abstract}

This paper provides an extensive analysis of the performance of the EfficientNet image classifiers with several recent training procedures, in particular one that corrects the discrepancy between train and test images~\cite{Touvron2019FixRes}. 
The resulting network, called FixEfficientNet, significantly outperforms the initial architecture with the same number of parameters. 
 
For instance, our FixEfficientNet-B0 trained without additional training data achieves 79.3\% top-1 accuracy on ImageNet with 5.3M parameters. This is a +0.5\% absolute improvement over the Noisy student EfficientNet-B0 trained with 300M unlabeled images. 
An EfficientNet-L2 pre-trained with weak supervision on 300M unlabeled images  
and further optimized with FixRes  
achieves {88.5\%} top-1 accuracy (top-5: {98.7\%}), 
which establishes the new state of the art for ImageNet with a single crop.

These improvements are thoroughly evaluated with cleaner protocols than the one usually employed for Imagenet, and particular we show that our improvement remains in the experimental setting of ImageNet-v2, that is less prone to overfitting, and with ImageNet Real Labels. In both cases we also establish the new state of the art. 
\end{abstract}

\input{introduction}

\input{related}
\input{experiments}

\section{Conclusion}
\label{sec:conclusion}

The "Fixing Resolution" is a method that improves the performance of any model. 
It is a method that is applied as a fine-tuning step after the conventional training, during a few epochs only, which makes it very flexible. 
It is easily integrated into any existing training pipeline.
In our paper we proposed a thorough evaluation of the combination of the current state-of-the-art models, namely EfficientNet, with this improved training method. 

We provide an open-source implementation of our method~\footnote{http://github.com/facebookresearch/FixRes}.

\bibliographystyle{IEEEbib}
\bibliography{egbib}

\end{document}

%% file: introduction.tex
\section{Introduction}\label{sec:introduction}

In order to obtain the best possible performance from Convolutional neural nets (CNNs), the training and testing data distributions should match.
However, in image recognition, data pre-processing procedures are often different for training and testing:
the most popular practice is to extract a rectangle with random coordinates from the image to artificially increase the amount of training data.
This \emph{Region of Classification} (RoC) is then resized to obtain an image, or crop, of a fixed size (in pixels) that is fed to the CNN.
At test time, the RoC is instead set to a square covering the central part of the image, which results in the extraction of a \emph{center crop}.
Thus, while the crops extracted at training and test time have the same size, they arise from different RoCs, which skews the data distribution seen by the CNN.

\begin{figure}[t]
\includegraphics[width=\linewidth]{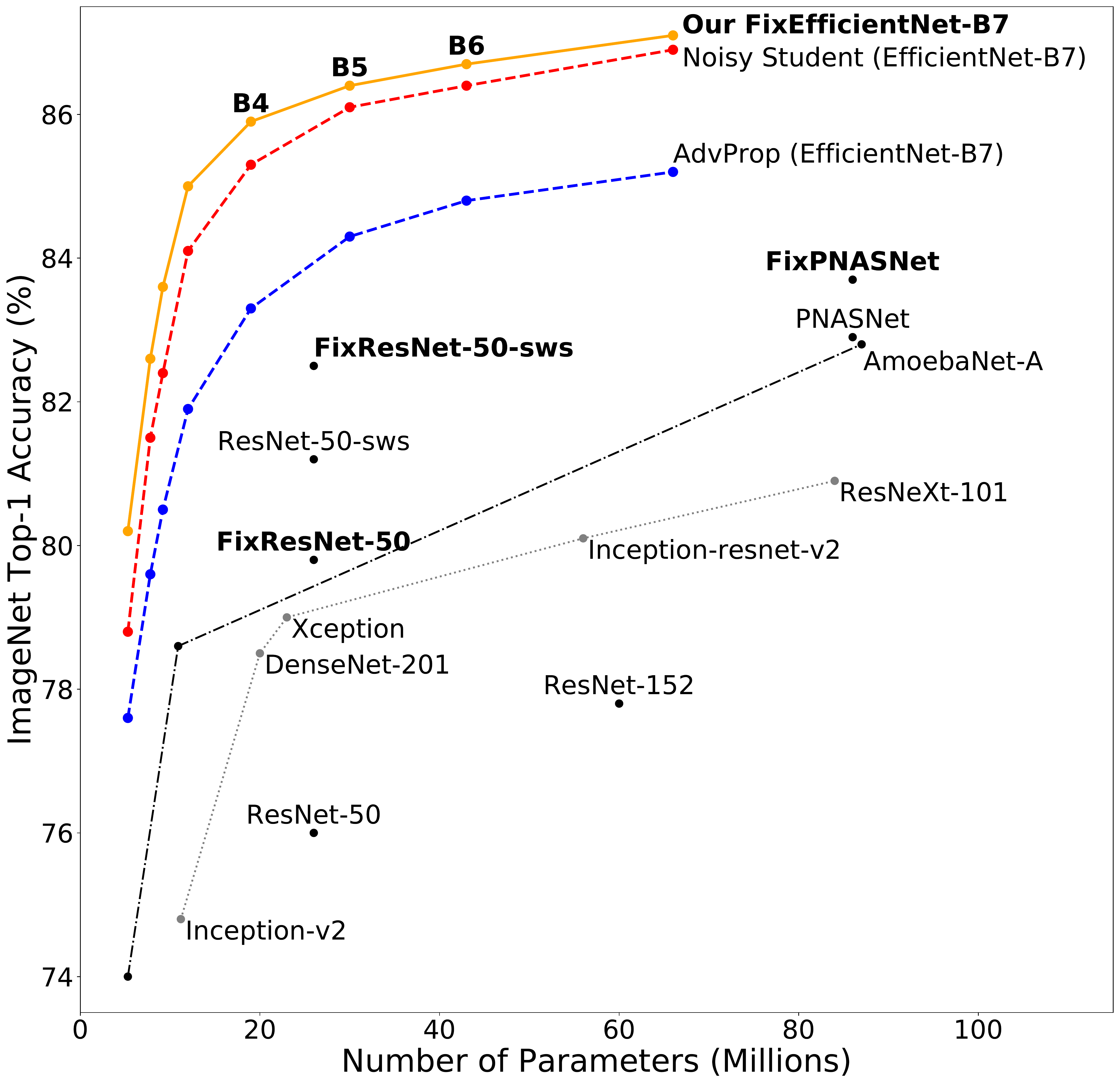}
\caption{\label{fig:FixEfficienNet}
Improvement brought by FixRes (in \textbf{bold}) to several popular architectures from the literature. 
Our FixEfficientNet (\textcolor{orange}{orange curve}) surpasses all EfficientNet models, including the models trained with Noisy student (\textcolor{red}{red curve}) and  adversarial examples (\textcolor{blue}{blue curve}).
The sws models are from~\cite{Yalniz2019BillionscaleSL}. 
Tables~\ref{tab:sota_extra_data} and~\ref{tab:sota} report results on larger models.
}
\end{figure}

Over the years, training and testing pre-processing procedures have evolved, but so far they have been optimized separately~\cite{Ekin2018AutoAugment}. 
Touvron et al. show~\cite{Touvron2019FixRes} that this separate optimization has a %
detrimental effect on the test-time performance of models.
They address this problem with the \emph{FixRes} method, which jointly optimizes the choice of resolutions and scales at training and test time, while keeping the same RoC sampling.

We apply this method to the recent EfficientNet~\cite{tan2019efficientnet} architecture, which offers an excellent compromise between number of parameters and accuracy. 
This evaluation paper shows that properly combining FixRes and EfficientNet further improves the state of the art~\cite{tan2019efficientnet}. 
Noticeably, 
\begin{itemize}
    \item We report the best performance without external data on ImageNet (top1: \textbf{85.7\%}); \\[-10pt]
    \item We report the best accuracy (top1: \textbf{88.5\%}) with external data on ImageNet, and with ImageNet with Reallabels~\cite{Beyer2020AreWD} ; \\[-10pt]
    \item We achieve state-of-the-art compromises between accuracy and number of parameters, see Figure~\ref{fig:FixEfficienNet};  \\[-10pt]
    \item We validate the significance of our results on the ImageNet-v2 test set, an improved evaluation setup that clearly separates the validation and test sets. FixEfficientNet achieves the best performance. 
\end{itemize}

This paper is organized as follows. In Section~\ref{sec:trainingupdates} we introduce the corrected training procedure for EfficientNet, that produces FixEfficientNet. Section~\ref{sec:experiments} analyzes our extensive evaluation and compare FixEfficientNet with the state of the art. Section~\ref{sec:conclusion} concludes the paper.

%% file: related.tex
\section{Training with FixRes: updates}\label{sec:related}
\label{sec:trainingupdates}
Recent research in image classification tends towards 
larger networks and higher resolution images~\cite{Yanping2018GPipe,mahajan2018exploring,Xie2019SelftrainingWN}.
For instance, the state-of-the-art in the ImageNet ILSVRC 2012 benchmark is currently held by the EfficientNet-L2~\cite{Xie2019SelftrainingWN} architecture with 480M parameters using 800$\times$800 images for training.
Similarly, the state-of-the-art model learned  from  scratch  is  currently EfficientNet-B8~\cite{Xie2019AdversarialEI} with 88M parameters using 672$\times$672 images  for training.
In this note, we focus on the EfficientNet architecture~\cite{tan2019efficientnet} due to its good accuracy/cost trade-off and its popularity.

\bigskip
\noindent \textbf{Data augmentation} is routinely employed at training time to improve model generalization and reduce overfitting. 
In this note, we use the same augmentation setup as in the original FixRes paper~\cite{Touvron2019FixRes}.
In addition, we have integrated label smoothing, which is orthogonal to the approach.
{\bf FixRes}
is a very simple fine-tuning that re-trains the classifier or a few top layers at the target resolution. 
Therefore, it has several advantages:
\begin{enumerate}
\item it is computationally cheap, the back-propagation is not performed on the whole network; \\ [-15pt]
\item it works with any CNN classification architecture and is complementary with the other tricks mentioned above;  \\ [-15pt]
\item it can be applied on a CNN that comes from a possibly non reproducible source. 
\end{enumerate}

%% file: experiments.tex
\section{Experiments}
\label{sec:experiments}

\newcommand{\bs}{B}

\label{sec:setup}

We experiment on the ImageNet-2012 benchmark~\cite{Russakovsky2015ImageNet12}, and report standard performance metrics (top-1 and top-5 accuracies) on a single image crop.

\subsection{Experimental Setting}

We focus on the EfficientNet~\cite{tan2019efficientnet} architectures. 
In the literature, wo versions provide the best performance:
EfficientNet trained with adversarial examples~\cite{Xie2019AdversarialEI}, and  
EfficientNet trained with Noisy student~\cite{Xie2019SelftrainingWN} pre-trained in a weakly-supervised fashion on 300 million unlabeled images.

We start from the EfficientNet models in rwightman's GitHub repository~\cite{pretrainedEffnet}. 
These models have been converted from the original Tensorflow to PyTorch. 

\bigskip \noindent
\textbf{Training. \quad}
We mostly follow the FixRes~\cite{Touvron2019FixRes} training protocol. 
The only difference is that we combine the FixRes data-augmentation with label smoothing during the fine-tuning.

\begin{table}
\centering 
{
\begin{minipage}{0.95\linewidth}
{\small
\centering
\caption{\label{tab:sota_extra_data}
  Results on \textbf{ImageNet with extra training data}. We start from pre-trained models~\cite{Xie2019SelftrainingWN} learned using 300M additional unlabeled images (single crop evaluation).
  See Section~\ref{sec:significance} about the significance of these results.
}
\smallskip
\begin{tabular}{|c|@{\ }r@{\quad}c|@{\ }c@{\quad}c@{\quad}c@{\ \ }|@{\ }c@{\quad}c@{\quad}c@{\ \ }|}
  \toprule
\multirow{2}{*}{\rotatebox{90}{Model\ \ \ }}  & \multirow{2}{*}{\rotatebox{90}{\#params}\ \ }  & \multirow{2}{*}{\rotatebox{90}{train res}} & \multicolumn{3}{c}{EfficientNet~\cite{Xie2019SelftrainingWN}} & \multicolumn{3}{c|}{FixEfficientNet} \\
 &  & 
 & test  & Top-1 &  Top-5  
 & test & Top-1 &  Top-5  \\ 
 ~ & & & res & (\%) & (\%) & res  & (\%) & (\%) \\
\midrule	
B0  & 5.3M  & 224  & 224 & 78.8 & 94.5 &320 & \textbf{80.2} & \textbf{95.4} \\
B1  & 7.8M  & 240  & 240 & 81.5 & 95.8 &384 & \textbf{82.6} & \textbf{96.5} \\
B2  & 9.2M  & 260  & 260 & 82.4 & 96.3 &420& \textbf{83.6} & \textbf{96.9} \\
B3  & 12M   & 300  & 300 & 84.1 & 96.9 &472 & \textbf{85.0} & \textbf{97.4} \\
B4  & 19M   & 380  & 380 & 85.3 & 97.5 &472 & \textbf{85.9} & \textbf{97.7} \\
B5  & 30M   & 456  & 456 & 86.1 & 97.8 &576 & \textbf{86.4} & \textbf{97.9} \\
B6  & 43M   & 528  & 528 & 86.4 & 97.9 &680 & \textbf{86.7} & \textbf{98.0} \\
B7  & 66M   & 600  & 600 & 86.9 & 98.1 & 632 & \textbf{87.1} & \textbf{98.2} \\
L2  & 480M  & 475  & 800 & 88.4 &  \textbf{98.7} & 600 & \textbf{88.5}& \textbf{98.7}\\
\bottomrule
\end{tabular}}
\end{minipage}
}
\end{table}

\begin{table}
\centering 
\begin{minipage}{0.95\linewidth}
\centering 
\caption{\label{tab:sota}
  Results on \textbf{ImageNet without external data} (single Crop evaluation).
  FixEfficientNet outperforms the previous EfficientNet AdvProp~\cite{Xie2019AdversarialEI} state of the art in this setup,
  see Section~\ref{sec:significance} for the significance of these results.
}
\smallskip 
{\small
\begin{tabular}{|c|@{\ }r@{\quad}c|@{\ }c@{\quad}c@{\quad}c@{\ \ }|@{\ }c@{\quad}c@{\quad}c@{\ \ }|}
  \toprule
\multirow{2}{*}{\rotatebox{90}{Model\ \ \ }}  & \multirow{2}{*}{\rotatebox{90}{\#params}\ \ }  & \multirow{2}{*}{\rotatebox{90}{train res}} & \multicolumn{3}{c}{EfficientNet~\cite{Xie2019AdversarialEI}} & \multicolumn{3}{c|}{FixEfficientNet} \\
 &  & 
 & test  & Top-1 &  Top-5  
 & test & Top-1 &  Top-5  \\ 
 ~ & & & res & (\%) & (\%) & res  & (\%) & (\%) \\
\midrule	
B0  & 5.3M  & 224  & 224 & 77.6 & 93.3 & 320 & \textbf{79.3} & \textbf{94.6} \\
B1  & 7.8M  & 240  & 240 & 79.6 & 94.3 & 384 & \textbf{81.3} & \textbf{95.7} \\
B2  & 9.2M  & 260  & 260 & 80.5 & 95.0 & 420 & \textbf{82.0} & \textbf{96.0} \\
B3  & 12M   & 300  & 300 & 81.9 & 95.6 & 472 & \textbf{83.0} & \textbf{96.4} \\
B4  & 19M   & 380  & 380 & 83.3 & 96.4 &512 & \textbf{84.0} & \textbf{97.0} \\
B5  & 30M   & 456  & 456 & 84.3 & 97.0 &576 & \textbf{84.7} & \textbf{97.2} \\
B6  & 43M   & 528  & 528 & 84.8 & 97.1 &576 & \textbf{84.9} & \textbf{97.3} \\
B7  & 66M   & 600  & 600 & 85.2 & 97.2 &632 & \textbf{85.3} & \textbf{97.4} \\
B8  & 87.4M & 672  & 672 & 85.5 & 97.3 & 800 & \textbf{85.7} & \textbf{97.6} \\

\bottomrule

\end{tabular}}
\end{minipage}
\end{table}

\begin{table}
\centering 
{
\begin{minipage}{0.95\linewidth}
{\small
\centering
\caption{\label{tab:sota_extra_data_real}
  Results on \textbf{ImageNet Real labels}~\cite{Beyer2020AreWD}.
}
\smallskip
\scalebox{0.8}{
\begin{tabular}{|c|cc|cc|cc|cc|}
  \toprule
\multirow{3}{*}{\rotatebox{90}{Model\ \ \ }}  & \multicolumn{4}{c}{No Extra-Training Data}
& \multicolumn{4}{c|}{Extra-Training Data} \\

& \multicolumn{2}{c}{EfficientNet~\cite{Xie2019SelftrainingWN}} & \multicolumn{2}{c}{FixEfficientNet} & \multicolumn{2}{c}{EfficientNet~\cite{Xie2019SelftrainingWN}} & \multicolumn{2}{c|}{FixEfficientNet}\\
 & Top-1 &  Top-5  & Top-1 &  Top-5 & Top-1 &  Top-5  & Top-1 &  Top-5  \\ 
 ~ & (\%) & (\%) & (\%) & (\%)& (\%) & (\%) & (\%) & (\%) \\
\midrule	
B0  & 83.7 & 95.8 & \textbf{85.8} & \textbf{96.8} & 84.5 & 96.4  & \textbf{86.5} & \textbf{97.3}  \\
B1  & 85.1 & 96.4  & \textbf{87.0} & \textbf{97.4}  & 86.7 & 97.2  & \textbf{88.1} & \textbf{98.0} \\
B2  & 86.0 & 96.8 & \textbf{87.7} & \textbf{97.6} & 87.3 & 97.6  & \textbf{88.8} & \textbf{98.2}  \\
B3  & 87.2 & 97.4 & \textbf{88.3} & \textbf{98.0} & 88.4 & 98.0  & \textbf{89.2} & \textbf{98.4}  \\
B4  & 88.3 & 97.9 & \textbf{89.2} & \textbf{98.3} & 89.4 & 98.4  & \textbf{89.8} & \textbf{98.5}  \\
B5  & 88.9 & 98.2 & \textbf{89.4} & \textbf{98.4} & 89.7 & 98.5  & \textbf{90.0} & \textbf{98.6}  \\
B6  & 89.3 & 98.3 & \textbf{89.6} & \textbf{98.4} & 89.8 & 98.5  & \textbf{90.1} & \textbf{98.6}  \\
B7  & 89.4 & 98.3 & \textbf{89.7} & \textbf{98.5} & 90.1 & 98.6  & \textbf{90.3} & \textbf{98.7}  \\
B8  & 89.6 & 98.3  & \textbf{90.0} & \textbf{98.6}& \_   & \_    & \_            & \_  \\
L2  & \_   & \_    & \_   & \_& 90.6 &  \textbf{98.8} & \textbf{90.9}& \textbf{98.8} \\
\bottomrule
\end{tabular}}}
\end{minipage}
}
\end{table}

\subsection{Comparison with the state of the art}

Table~\ref{tab:sota_extra_data} and Table~\ref{tab:sota} compare our results with those of the EfficientNet reported in the literature. 
All our FixEfficientNets outperform the corresponding EfficientNet (see Figure~\ref{fig:FixEfficienNet}). As a result and to the best of our knowledge, our FixEfficientNet-L2 surpasses all other results reported in the literature.
It achieves \textbf{88.5\%} Top-1 accuracy and \textbf{98.7\%} Top-5 accuracy on the ImageNet-2012 validation benchmark~\cite{Russakovsky2015ImageNet12}.

\bigskip \noindent
\textbf{Clean labels. \quad} 
In order to complement this evaluation, Table~\ref{tab:sota_extra_data_real}  present the results with the ImageNet clean labels proposed by Beyer et all.~\cite{Beyer2020AreWD}. 
With \textbf{90.9\%} Top-1 accuracy and \textbf{98.8\%} Top-5 accuracy FixEfficientNet-L2 surpasses all other results reported in the literature with this labels.

\subsection{Significance of the results}
\label{sec:significance}

Several runs of the same training incur variations of about 0.1 accuracy points on Imagenet due to random initialization and mini-batch sampling.
In general, since the Imagenet 2012 test set is not available, most works tune the hyper-parameters on the validation set, ie.  there is no distinction between validation and test set. This setting, while widely adopted, is not legitimate and can cause overfitting to go unnoticed.

EfficientNets employ Neural Architecture Search, which significantly enlarges the hyper-parameter space.   
Additionally, the ImageNet validation images were used to filter the images from the unlabelled set~\cite{Xie2019SelftrainingWN}. 
Therefore the pre-trained models may benefit from more overfitting on the validation set. 
We quantify this in the experiments presented below. 

Since we use pre-trained EfficientNet for our initialization, our results are comparable to those from the Noisy Student~\cite{Xie2019SelftrainingWN}, which uses the same degree of overfitting, but not directly with other semi-supervised approaches like that of Yalniz et al.~\cite{Yalniz2019BillionscaleSL}.

\begin{figure}
\begin{center}
\includegraphics[width=0.9\columnwidth]{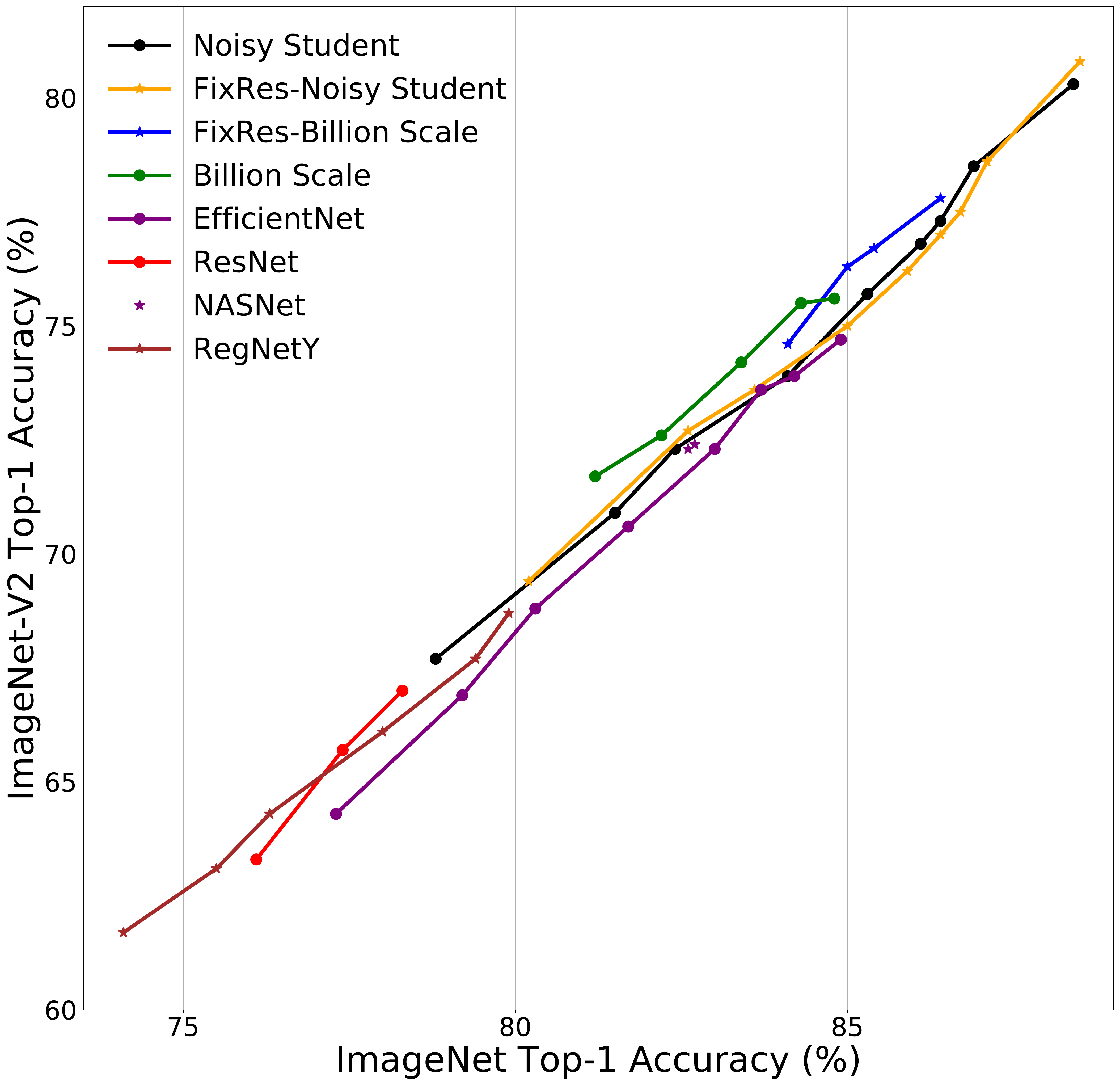}
\end{center}%
\caption{\label{fig:ImageNetv2_org}
Evidence of overfitting on Imagenet-val: We compare the results obtained on ImageNet (x-axis) and the results obtained on ImageNet-v2 (y-axis), without FixRes for different models~\cite{Zoph2018LearningTA,Liu2018PNAS,Xie2019SelftrainingWN,Radosavovic2020DesigningND,cubuk2019randaugment,Yalniz2019BillionscaleSL,He2016ResNet,Touvron2019FixRes}. 
For a given performance on Imagenet-val, overfitted models tend to have a lower performance on ImageNet-v2 and therefore are below the approaches that  generalize better. 
}
\end{figure}

\subsection{Evaluation on ImageNet-V2}
\label{sec:ImageNet_V2}

The ImageNet-V2~\cite{Recht2019DoIC} dataset was introduced to overcome the lack of a test split in the Imagenet dataset. 
ImageNet-V2 consists of 3 novel test sets that replace the ImageNet test set, which is no longer available. 
They were carefully designed to match the characteristics of the original test set.
One of these test sets, \textbf{Matched Frequency} is the closest to the ImageNet validation set. %
To ensure that observed improvements are not due to overfitting, we evaluate all our models on the Matched Frequency version of the ImageNet-v2~\cite{Recht2019DoIC} dataset.
We evaluate the other methods in the same way. 
We present the results in Tables~\ref{tab:sota_extra_data_v2} and \ref{tab:sota_v2}.

The original study of~\cite{Recht2019DoIC} shows that there is significant overfitting of various models to the Imagenet 2012 valuation set, but that it does not impact the relative order of the models.

\bigskip \noindent
\textbf{Quantifying the overfitting on Imagenet.} 
As mentioned earlier, 
several choices in the Noisy Student~\cite{Xie2019SelftrainingWN} method are prone to overfitting. 
We verify this hypothesis and quantify its extent by comparing the relative accuracy of this approach with another semi-supervised approach~\cite{Yalniz2019BillionscaleSL} both on ImageNet and ImageNet-V2~\cite{Recht2019DoIC}. 

Without overfitting, models performing similarly on Imagenet should also have  similar performances on ImageNet-V2~\cite{Recht2019DoIC}. However, for a comparable performance on ImageNet, when evaluating on ImageNet-V2, the Billion scale models of Yalniz \etal ~\cite{Yalniz2019BillionscaleSL} outperform  the EfficientNets from Noisy Student. 
For example, FixResNeXt-101 32x4d~\cite{Touvron2019FixRes} has the same performance as EfficientNet-B3~\cite{Xie2019SelftrainingWN} on ImageNet but on ImageNet-V2 FixResNeXt-101 32x4d~\cite{Touvron2019FixRes} is better (+0.7\% Top-1 accuracy). 

This shows that the EfficientNet Noisy student~\cite{Xie2019SelftrainingWN} tends to overfit and does not generalize as well as the (prior) semi-supervised work~\cite{Yalniz2019BillionscaleSL} or other works of the literature. 
Figure~\ref{fig:ImageNetv2_org} illustrates this effect. 
The FixRes fine-tuning procedure is neutral with respect to overfitting: overfitted models remain overfitted and conversely. 

\bigskip \noindent
\textbf{Comparison with the state of the art. \quad} 
Despite overfitting, EfficientNet remains very competitive on ImageNet-V2, as reported in Table~\ref{tab:=sota_ImageNet_V2}. 
Interestingly, the FixEfficientNet-L2 that we fine-tuned from EfficientNet establishes the new state of the art with additional data on this  benchmark. 

\begin{table}
\centering 
{\small
\begin{minipage}{0.95\linewidth}
\centering
\caption{\label{tab:sota_extra_data_v2}
  Results on ImageNet-V2~\cite{Recht2019DoIC} Matched Frequency with extra-training data. We start from pre-trained models~\cite{Xie2019SelftrainingWN} that have been learned using 300M additional unlabeled images (single crop evaluation).
  }
\smallskip
\begin{tabular}{|c|@{\ }r@{\quad}c|@{\ }c@{\quad}c@{\quad}c@{\ \ }|@{\ }c@{\quad}c@{\quad}c@{\ \ }|}
  \toprule
\multirow{2}{*}{\rotatebox{90}{Model\ \ \ }}  & \multirow{2}{*}{\rotatebox{90}{\#params}\ \ }  & \multirow{2}{*}{\rotatebox{90}{train res}} & \multicolumn{3}{c}{EfficientNet~\cite{Xie2019SelftrainingWN}} & \multicolumn{3}{c|}{FixEfficientNet} \\
 &  & 
 & test  & Top-1 &  Top-5  
 & test & Top-1 &  Top-5  \\ 
 ~ & & & res & (\%) & (\%) & res  & (\%) & (\%) \\
\midrule	

\midrule	
B0  & 5.3M  & 224  & 224 & 67.7 & 88.1 &320 & \textbf{69.4} & \textbf{89.6} \\
B1  & 7.8M  & 240  & 240 & 70.9 & 90.1 &384 & \textbf{72.7} & \textbf{91.4} \\
B2  & 9.2M  & 260  & 260 & 72.3 & 91.1 &420& \textbf{73.6} & \textbf{92.0} \\
B3  & 12M   & 300  & 300 & 73.9 & 91.9 &472 & \textbf{75.0} & \textbf{93.0} \\
B4  & 19M   & 380  & 380 & 75.7 & 93.1 &472 & \textbf{76.2} & \textbf{93.6} \\
B5  & 30M   & 456  & 456 & 76.8 & 93.6 &576 & \textbf{77.0} & \textbf{94.0} \\
B6  & 43M   & 528  & 528 & 77.3 & 93.9 &680 & \textbf{77.5} & \textbf{94.3} \\
B7  & 66M   & 600  & 600 & 78.5 & 94.4 & 632 & \textbf{78.6} & \textbf{94.7} \\
L2  & 480M  & 475  & 800 & 80.3 &  95.8 & 600 & \textbf{80.8}& \textbf{96.1}\\
\bottomrule
\end{tabular}
\end{minipage}
}
\end{table}

\begin{table}
\centering 
{\small
\begin{minipage}{0.95\linewidth}
\centering
\caption{\label{tab:sota_v2}
  Results on ImageNet-V2~\cite{Recht2019DoIC} Matched Frequency without external data (single Crop evaluation).
}
\smallskip 
\begin{tabular}{|c|@{\ }r@{\quad}c|@{\ }c@{\quad}c@{\quad}c@{\ \ }|@{\ }c@{\quad}c@{\quad}c@{\ \ }|}
  \toprule
\multirow{2}{*}{\rotatebox{90}{Model\ \ \ }}  & \multirow{2}{*}{\rotatebox{90}{\#params}\ \ }  & \multirow{2}{*}{\rotatebox{90}{train res}} & \multicolumn{3}{c}{EfficientNet~\cite{Xie2019AdversarialEI}} & \multicolumn{3}{c|}{FixEfficientNet} \\
 &  & 
 & test  & Top-1 &  Top-5  
 & test & Top-1 &  Top-5  \\ 
 ~ & & & res & (\%) & (\%) & res  & (\%) & (\%) \\
\midrule	
B0  & 5.3M  & 224  & 224 & 65.5 & 85.6 & 320 & \textbf{67.8} & \textbf{87.9} \\
B1  & 7.8M  & 240  & 240 & 67.5 & 87.8 & 384 & \textbf{70.1} & \textbf{89.6} \\
B2  & 9.2M  & 260  & 260 & 68.9 & 88.4 & 420 & \textbf{70.8} & \textbf{90.2} \\
B3  & 12M   & 300  & 300 & 70.9 & 89.4 & 472 & \textbf{72.7} & \textbf{90.9} \\
B4  & 19M   & 380  & 380 & 72.9 & 91.0 &512 & \textbf{73.9} & \textbf{91.8} \\
B5  & 30M   & 456  & 456 & 74.6 & 92.0 &576 & \textbf{75.1} & \textbf{92.4} \\
B6  & 43M   & 528  & 528 & 75.4 & 92.4 &576 & \textbf{75.4} & \textbf{92.6} \\
B7  & 66M   & 600  & 600 & \textbf{76.1} & 93.0 &632 & 75.8 & \textbf{93.2} \\
B8  & 87.4M & 672  & 672 & \textbf{76.1} & 92.7 & 800 & 75.9 & \textbf{93.0} \\

\bottomrule

\end{tabular}
\end{minipage}
}
\end{table}

\begin{table}
    \caption{\label{tab:=sota_ImageNet_V2}
     Performance comparison and state of the art on ImageNet-v2, single crop with external data, sorted by top-1 accuracy.  
     NS: Noisy Student~\cite{Xie2019SelftrainingWN}. BS: Billion-scale~\cite{Yalniz2019BillionscaleSL}. 
    }
    \smallskip
    
    \centering
    {\small
    \begin{tabular}{|l@{}r|cc|}
      \toprule
     Model & size \ &  Top-1 (\%)   &Top-5 (\%) \\
    \midrule

    EfficientNet-B0~NS~\cite{Xie2019SelftrainingWN}  & 5.3M & 67.7  & 88.1 \\
    FixEfficientNet-B0    & 5.3M & 69.4  & 89.6 \\
    
    EfficientNet-B1~NS~\cite{Xie2019SelftrainingWN}   & 7.8M & 70.9   & 90.1 \\
    ResNet50~BS~\cite{Yalniz2019BillionscaleSL}  &   25.6M   & 71.7  &  90.5 \\
    EfficientNet-B2~NS~\cite{Xie2019SelftrainingWN}  & 9.1M & 72.3   & 91.1 \\
    ResNeXt-50 32x4d~BS~\cite{Yalniz2019BillionscaleSL}  &   25.1M   & 72.6  &  90.9 \\
    FixEfficientNet-B1  &  7.8M & 72.7   & 91.4 \\
    FixEfficientNet-B2 & 9.1M & 73.6   & 92.0 \\
    
    EfficientNet-B3~NS~\cite{Xie2019SelftrainingWN}  & 12.2M & 73.9   & 91.9 \\
    ResNeXt-101 32x4d~BS~\cite{Yalniz2019BillionscaleSL}  &   42.0M   & 74.2  &  92.0 \\
    FixResNeXt-101 32x4d~\cite{Touvron2019FixRes}   &   42.0M   & 74.6  &  92.7 \\
    
    FixEfficientNet-B3  & 12.2M & 75.0   & 93.0 \\
    ResNeXt-101 32x8d~BS~\cite{Yalniz2019BillionscaleSL}  &   88.0M   & 75.5  &  92.8 \\
    ResNeXt-101 32x16d~BS~\cite{Yalniz2019BillionscaleSL}  &   193.0M   & 75.6  &  93.3 \\
    
    EfficientNet-B4~NS~\cite{Xie2019SelftrainingWN} & 19.3M & 75.7   & 93.1 \\
    FixEfficientNet-B4  & 19.3M & 76.2   & 93.6 \\
    
    FixResNeXt-101 32x8d~\cite{Touvron2019FixRes}    &   88.0M &   76.3   &  93.4 \\
    
    FixResNeXt-101 32x16d~\cite{Touvron2019FixRes}  &   193.0M   &  76.7  &  93.4 \\
    
    EfficientNet-B5~NS~\cite{Xie2019SelftrainingWN}&  30.4M & 76.8   & 93.6 \\
    FixEfficientNet-B5 & 30.4M & 77.0   & 94.0 \\
    
    EfficientNet-B6~NS~\cite{Xie2019SelftrainingWN}&  43.0M & 77.3   & 93.9 \\
    FixEfficientNet-B6  & 43.0M & 77.5   & 94.3 \\
    
    FixResNeXt-101 32x48d~\cite{Touvron2019FixRes}  &   829.0M &   77.8   &  93.9 \\
        
    EfficientNet-B7~NS~\cite{Xie2019SelftrainingWN} & 66.4M & 78.5   & 94.4 \\
    FixEfficientNet-B7 & 66.4M & 78.6   & 94.7 \\
    
    EfficientNet-L2~NS~\cite{Xie2019SelftrainingWN} & 480.3M & 80.5   & 95.7 \\
    FixEfficientNet-L2 &  480.3M & \textbf{80.8}   & \textbf{96.1} \\
    
      \bottomrule
    \end{tabular}}
    \end{table}